\documentclass[10pt,twocolumn,letterpaper]{article}

\usepackage{amssymb}
\usepackage{afterpage}
\usepackage{array}
\usepackage{makecell}
\usepackage{booktabs} 
\usepackage{multirow}
\usepackage{microtype}
\usepackage{algorithm}
\usepackage{algorithmic}
\usepackage{amsthm,amsmath,lipsum}
\usepackage{mathrsfs}

\usepackage[pagenumbers]{cvpr} 

\usepackage[dvipsnames]{xcolor}

\definecolor{cvprblue}{rgb}{0.21,0.49,0.74}
\usepackage[pagebackref,breaklinks,colorlinks,citecolor=cvprblue]{hyperref}

\def\paperID{6048} 
\def\confName{CVPR}
\def\confYear{2024}

\title{Symmetrical Bidirectional Knowledge Alignment for Zero-Shot Sketch-Based Image Retrieval}

\author{Decheng Liu\textsuperscript{1}, Xu Luo\textsuperscript{1}, Chunlei Peng\textsuperscript{1\thanks{Corresponding Author}}, Nannan Wang\textsuperscript{1}, Ruimin Hu\textsuperscript{1}, Xinbo Gao\textsuperscript{2}\\
\textsuperscript{1}Xidian University, Xi'an, China\\
\textsuperscript{2}Chongqing University of Posts and Telecommunications, Chongqing, China\\
}

\begin{document}
\maketitle
\begin{abstract}
This paper studies the problem of zero-shot sketch-based image retrieval (ZS-SBIR), which aims to use sketches from unseen categories as queries to match the images of the same category. Due to the large cross-modality discrepancy, ZS-SBIR is still a challenging task and mimics realistic zero-shot scenarios. The key is to leverage transferable knowledge from the pre-trained model to improve generalizability. 
Existing researchers often utilize the simple fine-tuning training strategy or knowledge distillation from a teacher model with fixed parameters, lacking efficient bidirectional knowledge alignment between student and teacher models simultaneously for better generalization. In this paper, we propose a novel Symmetrical Bidirectional Knowledge Alignment for zero-shot sketch-based image retrieval (SBKA). 
The symmetrical bidirectional knowledge alignment learning framework is designed to effectively learn mutual rich discriminative information between teacher and student models to achieve the goal of knowledge alignment. Instead of the former one-to-one cross-modality matching in the testing stage, a one-to-many cluster cross-modality matching method is proposed to leverage the inherent relationship of intra-class images to reduce the adverse effects of the existing modality gap. Experiments on several representative ZS-SBIR datasets (Sketchy Ext dataset, TU-Berlin Ext dataset and QuickDraw Ext dataset) prove the proposed algorithm can achieve superior performance compared with state-of-the-art methods.
\end{abstract}    
\section{Introduction}
\label{sec:intro}

Sketch-based image retrieval (SBIR) aims to recognize query sketch classes from large-scale gallery photos. 
These sketches, which are generated by hand, contain abstract lines and lack rich texture information compared with photo images.
It is difficult to conduct SBIR tasks due to the existing image modality gaps.
With the development of deep learning techniques, conventional SBIR methods have achieved satisfactory performance when the training and testing sets belong to the same categories.
Deep learning-based classification models can extract strong discriminative features for seen categories \cite{dutta2019semantically, tian2022tvt, lin2023zero, pmlr-v139-zhou21e}.
Considering the data scarcity problem of sketch collection, the zero-shot sketch-based image retrieval task is introduced to mimic real-world scenarios under zero-shot settings.
Thus, it is still an important and challenging problem in real-world applications.

Nowadays, expensive and time-consuming data collection and labeling make it essential to explore strong discriminative representation under zero-shot settings, even for different image modalities in real-world scenarios \cite{liu2023modality, liu2023fedforgery}. 
Early SBIR methods \cite{eitz2010sketch} utilized hand-crafted features to extract modality-invariant information to decrease the inevitable modality gap.
However, the key properties of discriminative features are not only cross-modality ability but also good generalization for unseen categories under zero-shot settings.
Benefiting from the strong generalization of the deep learning model, existing zero-shot sketch-based image retrieval methods are roughly grouped into two classes: convolutional network-based methods \cite{yelamarthi2018zero, dey2019doodle} and transformer-based methods \cite{tian2022tvt, lin2023zero}.
Although these works have achieved remarkable development, they often significantly depend on the discriminability of the refined backbone.
Limited works explore the design of a suitable bidirectional knowledge alignment for strong generalization and focus on decreasing the intra-class variance of gallery photos when testing.

The key to the ZS-SBIR task is to sufficiently learn discriminative knowledge from the fine original domain (e.g., rich discriminative knowledge from ImageNet or other large-scale models).
Most previous works only chose to fine-tune classification networks with parameters pre-trained on the ImageNet dataset and then designed suitable loss functions to capture cross-modality features.
Unlike a simple fine-tuning strategy, \cite{liu2019semantic} leveraged semantic information fusion with ImageNet pre-trained model and WordNet to achieve knowledge preservation. \cite{dong2023adapt} utilized the image-text large foundation model to align the learned image embedding and achieve better knowledge transfer.
However, these mentioned methods only consider unidirectional knowledge distillation from fine original domains containing fine and rich information, which cannot provide bidirectional knowledge alignment learning for better generalization.
Besides, previous ZS-SBIR works only focus on extracting robust and strong discriminative features for query sketches in the training stage, ignoring the decrease in intra-class variance in gallery photos in the testing stage.
\cite{wu2023distribution} designed a Cluster-then-Retrieve method to select the cluster centroids to replace gallery photos for retrieval results. However, the simple k-means clustering algorithm does not work well for diverse high-dimensional data with complex boundaries.
This still proves the gallery photos processing strategy has the potential to achieve satisfactory performance for the ZS-SBIR task.

To address these mentioned problems, we propose a novel symmetrical bidirectional knowledge alignment (SBKA) for zero-shot sketch-based image retrieval.
To avoid over-fitting and modality discrepancy problems in the ZS-SBIR task, the key issue is to learn modality-invariant discriminative knowledge from the fine original domain sufficiently.
Unlike from the former simple fine-tuning strategy or unidirectional knowledge distillation from the teacher model with fixed parameters, the proposed SBKA introduces novel bidirectional knowledge alignment learning to update parameters in both teacher and student models.
The motivation behind this is that these teacher and student models have different learning abilities in different domains, and the bidirectional knowledge communication when training makes it easier for better knowledge alignment and avoiding catastrophic forgetting \cite{liu2019semantic, han2018co}.
Additionally, the designed one-to-many cluster cross-modality matching also provides a novel perspective to decrease the intra-class variance of gallery photos for boosting retrieval performance.
Here the calculated similarities of gallery photos are aggregated through the intra-class relationship, potentially alleviating the adverse effects of limited cluster performance.

\par The main contributions of our paper are summarized as follows:
\begin{enumerate}
\item We propose a novel symmetrical bidirectional knowledge alignment learning framework for the ZS-SBIR task, which enables both teacher and student models to learn mutual discriminative information bidirectionally for better knowledge alignment.
\item The one-to-many cluster cross-modality matching algorithm is designed to efficiently mine the relationship of intra-class gallery photos, which provides a novel perspective to boost cross-modality matching performance.
\item Experimental results demonstrate the superior performance of the proposed method SBKA compared with state-of-the-art zero-shot sketch-based image retrieval algorithms on representative public Sketchy Ext, TU-Berlin Ext, and QuickDraw Ext datasets.

\end{enumerate}


\begin{figure*}[h!]
\centering
\includegraphics[width=0.68\linewidth]{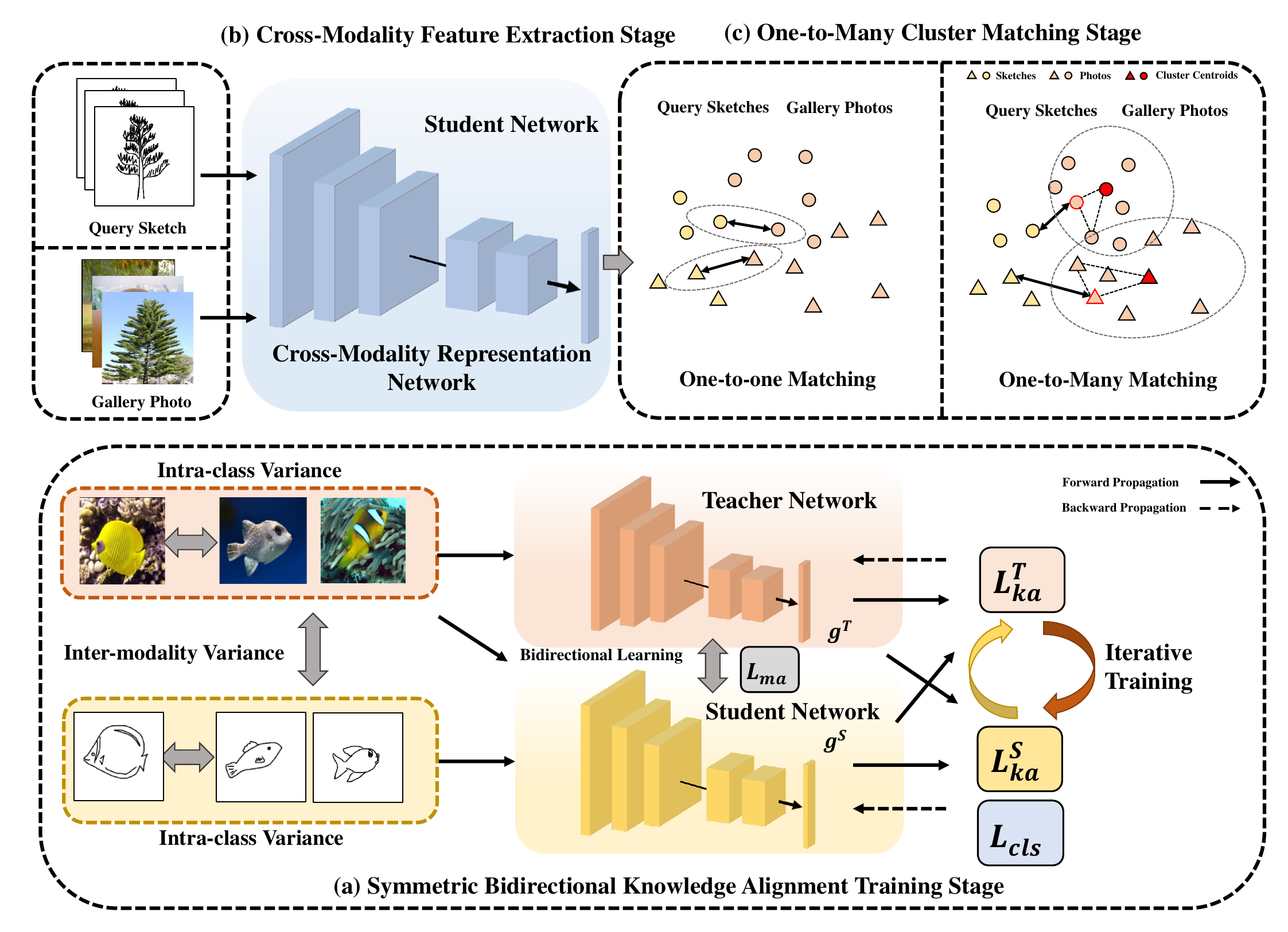} 
\caption{Overview of the proposed symmetrical bidirectional knowledge alignment framework for zero-shot sketch-based image retrieval.
The proposed method consists of three stages: (a) symmetrical bidirectional knowledge alignment training stage; (b) cross-modality feature extraction stage; (c) one-to-many cluster matching stage. 
}
\label{fig_framework}
\end{figure*}

\section{Related Work}
\label{sec:formatting}


\subsection{Zero-shot Sketch-based Image Retrieval}

Early SBIR methods used handcrafted features to replace sketch vectors for cross-modal photo image retrieval, such as \cite{eitz2010evaluation}, \cite{eitz2010sketch}, \cite{saavedra2015sketch}.
With the development of deep learning technologies, 
more and more works have begun to leverage deep networks to improve the generalization of unseen categories.
\cite{shen2018zero} first introduced the ZS-SBIR task and proposed an end-to-end three-way network architecture hashing model. 
Nowadays, ZS-SBIR works are roughly categorized into two types: CNN-based methods and transformer-based methods. 
\cite{yelamarthi2018zero} generated the missing additional information in sketches and further employed conditional generative models to stochastically complete sketches. 
\cite{dutta2019semantically} utilized an adversarial training framework to map sketches and photos into a shared feature space for retrieval. \cite{dey2019doodle} leveraged knowledge distillation and graph attention networks, which force the model to be trained with sketches drawn by users as new semantic information. 
\cite{liu2019semantic} utilized domain adaptation from a pre-trained CNN to boost the performance of the model.
With the development of the transformer network, recent researchers, as in \cite{tian2022tvt}, have leveraged Triplet-ViT to model global representation in the multi-modal hypersphere.
\cite{lin2023zero}  attempted to explain the matching process by using a self-attention module with a learnable tokenization. 
\cite{sain2023clip} adapted to the ZS-SBIR task by fine-tuning the CLIP model. During the testing phase, it employed prompt learning to introduce additional prompts as auxiliary information for sketch-based image retrieval. 

The above-mentioned works focus on training stronger backbone networks for extracting discriminative features, which regard the ZS-SBIR task as the classification task.
However, considering the specific properties of the sketch images, reducing intra-class variance in the gallery photos when testing provides a novel perspective to boost retrieval performance.

\subsection{Knowledge Distillation}

The knowledge distillation algorithm \cite{hinton2015distilling} was first proposed to address the over-fitting problem by condensing knowledge from a large teacher model into a compact student model.
Nowadays, the teacher-student structure is effectively and extensively utilized for various objectives in knowledge distillation. \cite{xie2020self} attempted to use the predictions of the teacher model as pseudo-labels for training a larger model. Through enhanced data augmentation, the student model's performance surpassed the teacher model. \cite{matiisen2019teacher} designed a curriculum learning framework in which the student model learned complex tasks.
Based on the student model's ability, the teacher model chose subtasks to train the student model further. 
\cite{ghiasi2021multi} used pseudo-labels generated by multiple independent teacher models, each designed for specific tasks, to train a single general student model.
In the ZS-SBIR field,  \cite{liu2019semantic} and \cite{tian2022tvt} typically utilized unidirectional knowledge distillation to enhance the student model's performance. 
However, the parameters of the teacher model are fixed, which cannot provide bidirectional knowledge alignment learning for better generalization.

\section{Methodology}

In this section, we propose a novel symmetrical bidirectional knowledge alignment framework for zero-shot sketch-based image retrieval (SBKA).  
Two main issues limit zero-shot cross-modality retrieval: large \textbf{inter-modality variance and intra-class variance}.
Figure \ref{fig_framework} shows the overall framework of the proposed SBKA algorithm, which consists of two key stages: 
The symmetrical bidirectional knowledge alignment training stage leverages the mutual interaction of teacher and student models to align stronger discriminative knowledge and boost generalizability;
The one-to-many cluster matching strategy is proposed to aggregate the discriminative information of gallery photos through mining intra-class relationships, which can effectively alleviate the effect of substantial intra-class variance.
In the following subsections, we will introduce the problem formulation first and then give a detailed explanation of our proposed method.

\subsection{Motivation}
In order to improve the generalization performance of the ZS-SBIR model, previous works \cite{liu2019semantic, tian2022tvt} mainly depended on a teacher model from the large-scale dataset (e.g., ImageNet), which was always finely pre-trained and then frozen for subsequent knowledge distillation.
This is because the finely trained original domain model contains rich discriminative information for alleviating the overfitting problem in zero-shot settings.
Thus, \emph{the key for the ZS-SBIR task is to sufficiently learn strong and robust knowledge from the source domain and transfer it to the target domain.}
The sketch-based image retrieval can be regarded as a cross-modality image recognition task, while the teacher model is only trained to adapt to the single source photo domain.
Due to the existing modality gap, the refined teacher model only transfers elaborate semantic information to the target hybrid domain, which contains both sketch and photo domains simultaneously.
The limited transfer ability indeed brings adverse effects to the cross-modality retrieval precision.
Most existing works typically choose the simple fine-tuning strategy or freeze the parameters of the refined teacher model, which works well only when transferring domain knowledge within the same modalities but hinders the knowledge alignment between different modality domains.
To address this issue, we aim to incorporate the symmetrical bidirectional knowledge alignment learning algorithm when training the target student network.
The refined teacher model is not frozen and will adapt parameters dynamically based on the supervised signal provided by the student model.
In other words, the proposed knowledge alignment learning is bidirectional. 
When the teacher model continually updates parameters dynamically, it can align knowledge and better adapt to both sketch and photo modalities.
Furthermore, it will provide a more suitable supervised signal to the target student model to boost performance. \par

\subsection{Problem Formulation}
Given a sketch-based image retrieval dataset denoted as $D = \left\{ {(x_i^{\bmod },{y_i})} \right\}_{i = 1}^M,\bmod  \in \{ S,P\} $, where $x_i$ means the $i$-th sample, $y_i$ is the corresponding hard label.
$S$ and $P$ represent the sketch modality and the photo modality, respectively.
As illustrated in the previous section, the goal of ZS-SBIR is to train a strong discriminative representation model on the source dataset $D_s$, and evaluate it on the target dataset $D_t$. 
The zero-shot setting means these $D_s$ and $D_t$ datasets contain different classes.
In the testing stage, the sketch $x_i^S$ is chosen as the query to retrieval in the gallery photos $x_j^P$.
Thus, the key is to learn a cross-modality representation ${F_\theta }( \cdot )$ with good generalization to avoid the overfitting problem.

\subsection{Symmetrical Bidirectional Knowledge Alignment Algorithm}
To effectively transfer discriminative knowledge well between the source teacher model and the target student model, we design the symmetrical bidirectional knowledge alignment learning algorithm to train the cross-modality representation network (as shown in Fig. \ref{fig_framework} (a)).
Instead of freezing all the parameters of the teacher model, we enable both teacher and student models to conduct bidirectional knowledge communication iteratively. 
It is noted both the student and teacher models utilize the same network structure for convenience.
Here we choose the ConvNeXtV2 \cite{woo2023convnext} network as the backbone network.
For each input sketch or photo ${x_i}$, the output logits can be denoted as 
${f_i} = {F_\theta }({x_i})$.
The predictive probability in the student representation network can be calculated by a softmax function, where the ${p_i}$ is denoted as:
\begin{equation}
{p_i} = \frac{{\exp ({f_i})}}{{\sum\nolimits_{j = 1}^K {\exp ({f_j})} }}
\label{softmax},
\end{equation}
where $K$ is the number of classes in the training set. The cross-entropy between the hard label and the prediction probability ${p_i}$ is defined as follows:
\begin{equation}
{L_{cls}} =  - \sum\limits_{i = 1}^N {\sum\limits_{k = 1}^K {{y_{i,k}}} } \log ({p_i}{\rm{)}},
\end{equation}
where ${N}$ denotes the number of samples in the training set.
Additionally, we leverage the Kullback-Leibler (KL) divergence constraint on sketch features and photo features to help decrease inter-modality variance.
Here, we align sketch and photo feature distribution with standard Gaussian distribution $ e^G \sim \mathcal{N}(0,1)$ for convenience.
$e_i^S$ and $e_i^P$ mean the extracted feature when inputting sketches and photos, respectively.
The designed inter-modality alignment constraint is denoted as:
\begin{equation}
{L_{ma}} = \sum\limits_{i = 1}^{{N_S}} {\sum\limits_{k = 1}^K {e_{_{i,k}}^S} } \log (\frac{{e_{_{i,k}}^S}}{{e_{_{i,k}}^G}}) + \sum\limits_{i = 1}^{{N_P}} {\sum\limits_{k = 1}^K {e_{_{i,k}}^P} } \log (\frac{{e_{_{i,k}}^P}}{{e_{_{i,k}}^G}}).
\end{equation}

For bidirectional knowledge alignment, we utilize the outputs from the teacher and student models as soft labels for training them alternately.
Inspired by \cite{liu2019semantic}, we additionally integrate the semantic similarity information extracted from the pre-trained WordNet \cite{miller1995wordnet} model, denoted as $a_i$, for improving performance.
Therefore, the supervised signal from the teacher model ${F_{{\theta _T}}}$ is set as the soft label for training the student model, which is represented as $g_i^T = \Phi (f_i^T + \lambda {a_i})$.
On the other hand, the supervised signal from the student model ${F_{{\theta _S}}}$ is set as the soft label for alternately training the teacher model, which is represented as $g_i^S = \Phi (f_i^S + \lambda {a_i})$.
Note that $\theta _S$ and $\theta _T$ mean parameters of the ImageNet label classifier \cite{liu2019semantic}, which contains most of the same parameters as the mentioned representation model, except for the auxiliary linear layer.
$\Phi ( \cdot )$ denotes the softmax function, the same as in Eq. \ref{softmax}.
In the target student network training stage, the teacher model needs to provide a soft label for learning mutual information. The bidirectional knowledge alignment constraint of the student model is defined as follows:
\begin{equation}
L_{_{ka}}^S =  - \sum\limits_{i = 1}^N {\sum\limits_{k = 1}^K {g_{_{i,k}}^T} } \log (p_{_i}^S{\rm{)}}.
\end{equation}

Therefore, the overall training objective of the student model can be summarized as follows:
\begin{equation}
L^S = L_{cls} + L_{ka}^S + \lambda L_{ma},
\label{equation5}
\end{equation}
where the hyperparameter $\lambda$ can balance the aforementioned loss terms. The detailed analysis of these parameters is discussed in the subsequent section.


Similarly, the student model provides a soft label for learning mutual information when training the teacher model. The bidirectional knowledge alignment constraint of the teacher model is defined as follows:
\begin{equation}
L_{_{ka}}^T =  - \sum\limits_{i = 1}^N {\sum\limits_{k = 1}^K {g_{_{i,k}}^S} } \log (p_{_i}^T{\rm{)}}.
\label{equation6}
\end{equation}
With the above loss functions determined, we train the SBKA representation model by alternately updating the student and teacher models.
Symmetrical bidirectional knowledge alignment is achieved by alternately updating parameters in both the teacher and student models.
The detailed training procedure is outlined in Algorithm \ref{alg:algorithm}. 

\begin{algorithm}[tb]
\caption{Training process of SBKA.}
\label{alg:algorithm}
\textbf{Input}: $x_i$, hard label $y_i$, batchsize $N$, learning rate $\alpha$. \\
\textbf{Parameter}: Teacher model parameters $\theta_T$, student model parameters $\theta_S$.\\
\textbf{Output}: {$\theta_S$, $\theta_T$}.
\begin{algorithmic}[1] 
\STATE Let $epoch=t$.
\WHILE{$epoch <$ max iterations}
\STATE Compute the loss $L^S \leftarrow \text{Eq. \ref{equation5}}$.
\STATE Compute the loss $L_{ka}^T \leftarrow \text{Eq. \ref{equation6}}$.
\STATE $\theta_S \mathrel{\mathop{\leftarrow}\limits^{+}} - \alpha \nabla_{\theta_S} L^S$.
\STATE $\theta_T \mathrel{\mathop{\leftarrow}\limits^{+}} - \alpha \nabla_{\theta_T} L_{ka}^T$.
\ENDWHILE
\STATE \textbf{return} {$\theta_S$, $\theta_T$}.
\end{algorithmic}
\end{algorithm}

\subsection{One-to-Many Cluster Cross-modality Matching Method}
There are both large inter-modality variance and intra-class variance in the ZS-SBIR dataset.
Most existing works focus on decreasing the obvious modality gap, ignoring the need to address intra-class variance.
Unlike previous one-to-one matching strategies (as shown in Fig. \ref{fig_framework} (c)) in the testing retrieval stage, the proposed one-to-many cluster matching algorithm can leverage the inherent intra-class relationship, offering a reasonable and novel perspective for boosting performance.
Naturally, we utilize the relationship of cluster centroids, which contain the most discriminative information.
Thus, the fundamental concept of one-to-many cluster matching is to directly select the cluster centroid of each class to replace the original gallery photos.
However, the preprocessing cluster algorithm cannot guarantee that each sample is correctly matched, which may lead to inaccurate proxies of gallery photos.
To address this issue, we propose the one-to-many cluster matching algorithm to effectively leverage relationships in the intra-class set with a score fusion strategy, rather than through direct replacement.

Given the query sketch and gallery photos, we aim to leverage the inherent relationship of gallery photos to decrease intra-class variance.
For each input sketch or photo ${x_i}$, the output feature can be denoted as 
${e_i} = {E_\theta }({x_i})$.
The Gaussian Mixture Model (GMM) is utilized to model the probability density function
of gallery photos features $e_i^P$ as follows:
\begin{equation}
p(e_i^P) = \sum\limits_{k = 1}^K {{w_k}N(e_i^P|{\mu _k},{C_k})} ,
\end{equation}
where $\sum {{w_k}}  = 1,{w_k} \in [0,1]$, the $i$-th component is characterized by normal Gaussian distributions with weights $w_k$, means ${\mu _k}$ and covariance matrices $C_k$.
This problem can be solved using the EM algorithm \cite{yu2011solving}.
The E-step and the M-step are iteratively repeated until convergence is reached.
The mean feature vector of each class in the gallery photo set is regarded as the cluster centroid $c_j^S$ for the following processing.

Inspired by the Production Quantization algorithm \cite{jegou2010product}, we divide the original embedding space into several subspaces to decrease the need for computing memory.
Given a gallery photo feature $e_i^P$, we divide it into $M$ subspaces as $e_i^P = \{ e_{i,1}^P,e_{i,2}^P,...,e_{i,M}^P\} $.
Thus, we can calculate $K \times M$ cluster centroids for the gallery photos set.
In the retrieval stage, we fuse the inherent relationship to decrease intra-class variance for better discriminability.
Specifically, the fused dissimilarity score is calculated for the query sketch $s_i$ and the gallery photo $p_j$ as follows:
\begin{equation}
{d_{i,j}}{\rm{ = }}\Psi (e_i^S,e_j^P) + \sum\limits_{m = 1}^M {\Psi (e_{i,m}^S,e_{j,m}^C)},
\end{equation}
where $e_{j,m}^C$ denotes the extracted feature of the corresponding cluster centroid in the $m$-th subspace.
Here we choose the Euclidean distance ${\left\|  \cdot  \right\|_2}$ as the distance function $\Psi ( \cdot )$.
Once the aggregated dissimilarity score is calculated, the retrieval results can be determined by sorting in descending order.


\begin{table*}[ht]
\centering
\caption{Comparison of results with state-of-the-art methods on public datasets.}
\label{table1}
\resizebox{\linewidth}{!}{
\begin{tabular}{ccccccccc}
    \toprule
    \multirow{2}{*}{Methods} & \multicolumn{2}{c}{Sketchy Ext} & \multicolumn{2}{c}{Sketchy Ext Split} & \multicolumn{2}{c}{TU-Berlin Ext} & \multicolumn{2}{c}{QuickDraw Ext} \\
    \cmidrule{2-3} \cmidrule{4-5} \cmidrule{6-7} \cmidrule{8-9}
    & mAP@all & Prec@100 & mAP@200 & Prec@200 & mAP@all & Prec@100 & mAP@all & Prec@200 \\
    \midrule
    CAAE (ECCV'18) & 0.196 & 0.284 & 0.156 & 0.260 & - & - & - & - \\
    CVAE (ECCV'18) & - & - & 0.225 & 0.333 & 0.005 & 0.001 & 0.003 & 0.003 \\
    SEM-PCYC (CVPR'19) & 0.349 & 0.463 & - & - & 0.297 & 0.426 & - & - \\
    DOODLE (CVPR'19) & - & - & 0.369 & - & 0.109 & - & 0.075 & 0.068 \\
    SAKE (ICCV'19) & 0.547 & 0.692 & 0.497 & 0.598 & 0.475 & 0.599 & 0.130 & 0.179 \\
    LCALE (AAAI'20) & 0.476 & 0.583 & - & - & - & - & - & - \\
    OCEAN (ICME'20) & 0.462 & 0.590 & - & - & 0.333 & 0.467 & - & - \\
    RPKD (ACMMM'21) & 0.613 & 0.723 & 0.502 & 0.598 & 0.486 & 0.612 & 0.143 & 0.218 \\
    DSN (IJCAI'21) & 0.583 & 0.704 & - & - & 0.481 & 0.591 & - & -  \\
    SBTKNet (PR'22) & 0.553 & 0.698 & 0.502 & 0.596 & 0.480 & 0.608 & 0.119 & 0.167 \\
    TVT (AAAI'22) & 0.648 & 0.796 & 0.531 & 0.618 & 0.484 &  \textbf{0.662} & 0.149 & \textbf{0.293} \\
    Sketch3T (CVPR'22) & 0.575 & - & - & - & 0.507 & - & - & - \\
    ZSE-SBIR (CVPR'23) & 0.736 & \textbf{0.808} & 0.525 & 0.624 & 0.569 & 0.637 & 0.145 & 0.216 \\
    \cmidrule{1-9}
    Ours & \textbf{0.799} & 0.764 & \textbf{0.618} & \textbf{0.666} & \textbf{0.635} & 0.605 & \textbf{0.198} & 0.186\\
    \bottomrule
\end{tabular}
}
\end{table*}

\section{Experiments}
To verify the effectiveness of our model, we initially conduct experiments on common ZS-SBIR datasets and compare it with current state-of-the-art (SOTA) methods. Subsequently, we carry out parameter analysis and ablation studies to demonstrate the impact of various key components on the model's performance. Finally, we provide retrieval examples and visualize image features to verify the discriminative information learned by our model in the zero-shot setting. This also illustrates the model's capacity for generalization.
\subsection{Experimental Settings}
\paragraph{Datasets.} \quad We verify our method on three generic datasets: Sketchy Ext \cite{sangkloy2016sketchy}, TU-Berin Ext \cite{eitz2010evaluation}, and QuickDraw Ext \cite{dey2019doodle}.
Sketchy Ext dataset consists of 125 categories with 75,471 sketches and 12,500 images. Researchers \cite{liu2019semantic} additionally collected 60,502 natural images, making a total of 73,002 images. 
TU-Berlin Ext dataset contains 20,000 sketches in 250 categories, with an additional 204,489 images, also collected by \cite{liu2019semantic}. 
QuickDraw Ext is currently the largest SBIR dataset, containing 330,000 sketches and 204,000 images across 110 categories. Unlike Sketchy Ext and TU-Berlin Ext, the sketches in QuickDraw Ext are highly abstract and stylistically diverse. As a result, cross-modal sketch retrieval on QuickDraw Ext poses a greater challenge. 
With the same protocol \cite{liu2019semantic}, we randomly divide Sketchy Ext into 100 categories as the training set and 25 categories as the testing set. Similarly, TU-Berlin Ext is divided into 200 training categories and 30 testing categories. 
We follow a similar protocol \cite{yelamarthi2018zero} to avoid the overlap of ImageNet categories with unseen categories in the Sketchy Ext dataset.
These 104 categories are randomly selected as the training set, while the rest of the categories are the testing set. For the QuickDraw Ext dataset, we adopt a methodology from \cite{dey2019doodle}, the 80 classes are selected as the training set, and the remaining classes are the testing set.

\paragraph{Evaluation Metric.}
Following recent ZS-SBIR work \cite{liu2019semantic, lin2023zero, dey2019doodle}, we select Mean Average Precision (mAP) and Precision as our evaluation metrics. Specifically, on the Sketchy Ext and TU-Berlin Ext datasets, we report mAP and Precision on the top 100 results (Prec@100). Consistent with other studies, for the Sketchy Ext Split and QuickDraw Ext, we report both mAP on the top 200 (mAP@200) and Prec@200, as well as mAP and Prec@200.

\paragraph{Implementation Details.} 
We conduct the proposed method using PyTorch on two NVIDIA RTX 3090 GPUs. For symmetrical bidirectional knowledge alignment, the student and teacher models adopt the same structure ConvNeXtV2-Tiny network.
To prevent early symmetrical bidirectional knowledge alignment from leading to model collapse, we freeze the gradients of the teacher model in the early training stage. 
The training epochs of designed symmetrical bidirectional knowledge alignment on Sketchy Ext, TU-Berlin Ext, and QuickDraw Ext datasets are set as 10, 6, and 4 respectively. 
The initial learning rate of the Adam optimizer for the student model is $1e–4$ and gradually decays to $1e–7$. 
For the Sketchy Ext and QuickDraw Ext datasets, the optimizer parameters of the teacher model are the same as the student model, while the initial learning rate of the teacher model for TU-Berlin Ext is $5e–5$ and decays to $7e–8$. In the one-to-many cluster matching phase, the number of subspaces is set to 4 on the Sketchy Ext and Sketchy Ext Split datasets, and the number of clusters is set as 25 and 21 respectively, which exactly equals the number of gallery photos categories. 
The number of subspaces is set to 2 on the TU-Berlin Ext and QuickDraw Ext datasets, and the number of clusters is set as 32 and 30 respectively.
\subsection{Comparison with SOTA Methods}

We compare our method with SOTA methods in the ZS-SBIR task, including CAAE and CVAE \cite{yelamarthi2018zero}, SEM-PCYC \cite{dutta2019semantically}, DOODLE \cite{dey2019doodle}, SAKE \cite{liu2019semantic}, LCALE \cite{lin2020learning}, OCEAN \cite{zhu2020ocean}, RPKD \cite{tian2021relationship}, DSN \cite{wang2021domain}, SBTKNet \cite{tursun2022efficient}, TVT \cite{tian2022tvt}, Sketch3T \cite{sain2022sketch3t}, and ZSE-SBIR \cite{lin2023zero}. Comparison experimental results are shown in Table \ref{table1}.
The comparison methods are roughly categorized into two types according to the backbones: CNN-based methods \cite{liu2019semantic, sain2022sketch3t, tursun2022efficient} 
 and transformer-based methods \cite{tian2022tvt, lin2023zero}.
Benefiting from its self-attention mechanism, these transformer-based methods generally outperform CNN-based models. 
In the Sketchy Ext dataset, these transformer-based methods can mostly achieve above 0.64 with the mAP metric, whereas CNN-based methods only achieve below 0.60.
Benefiting from the symmetrical bidirectional knowledge alignment learning strategy, the proposed CNN-based method still outperforms these transformer-based methods by at least $6\%$.

\begin{figure*}[ht]
\centering
\subcaptionbox{\hspace{-0.7cm}\label{Fig.sub.1}}{
    \includegraphics[width=4cm,height=3cm]{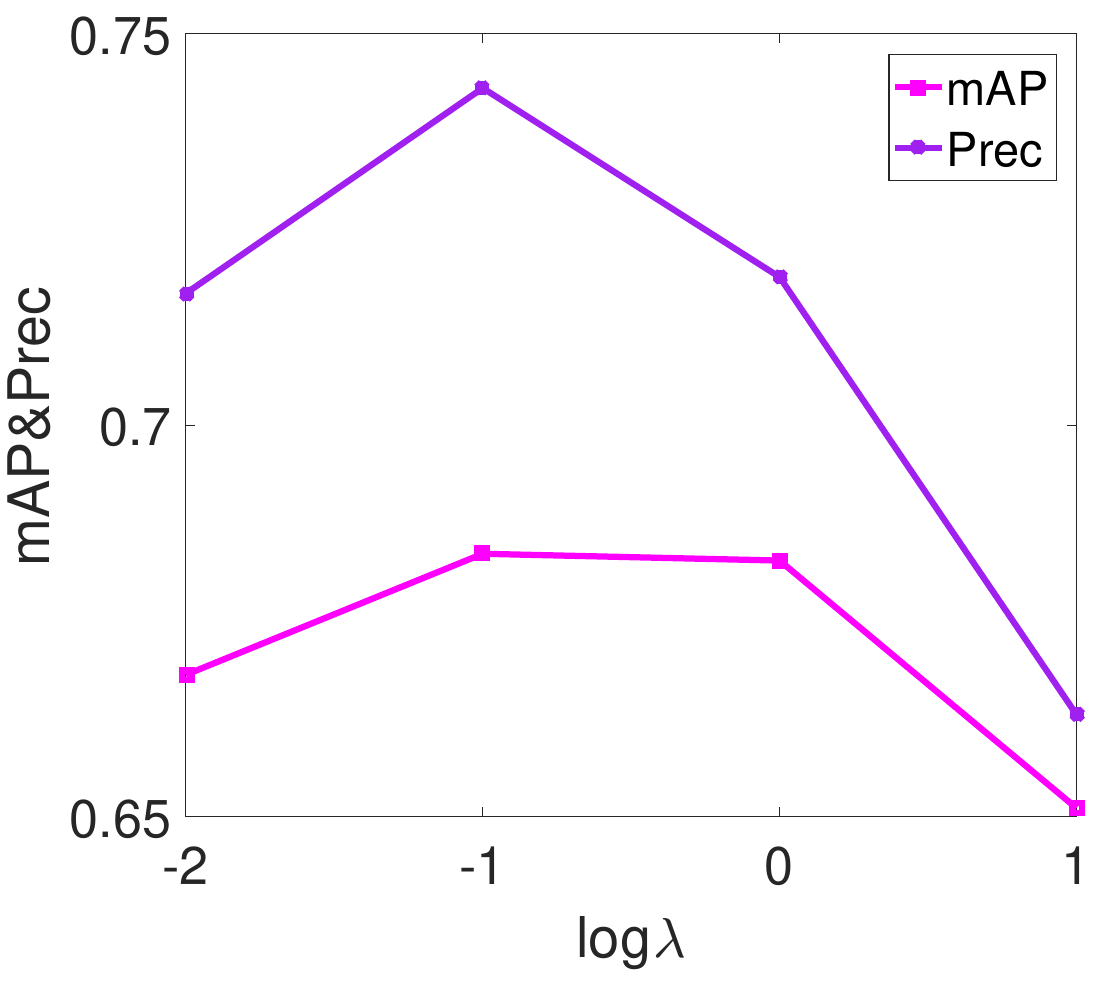}}
\subcaptionbox{\hspace{-0.7cm}\label{Fig.sub.2}}{
    \includegraphics[width=4cm,height=3cm]{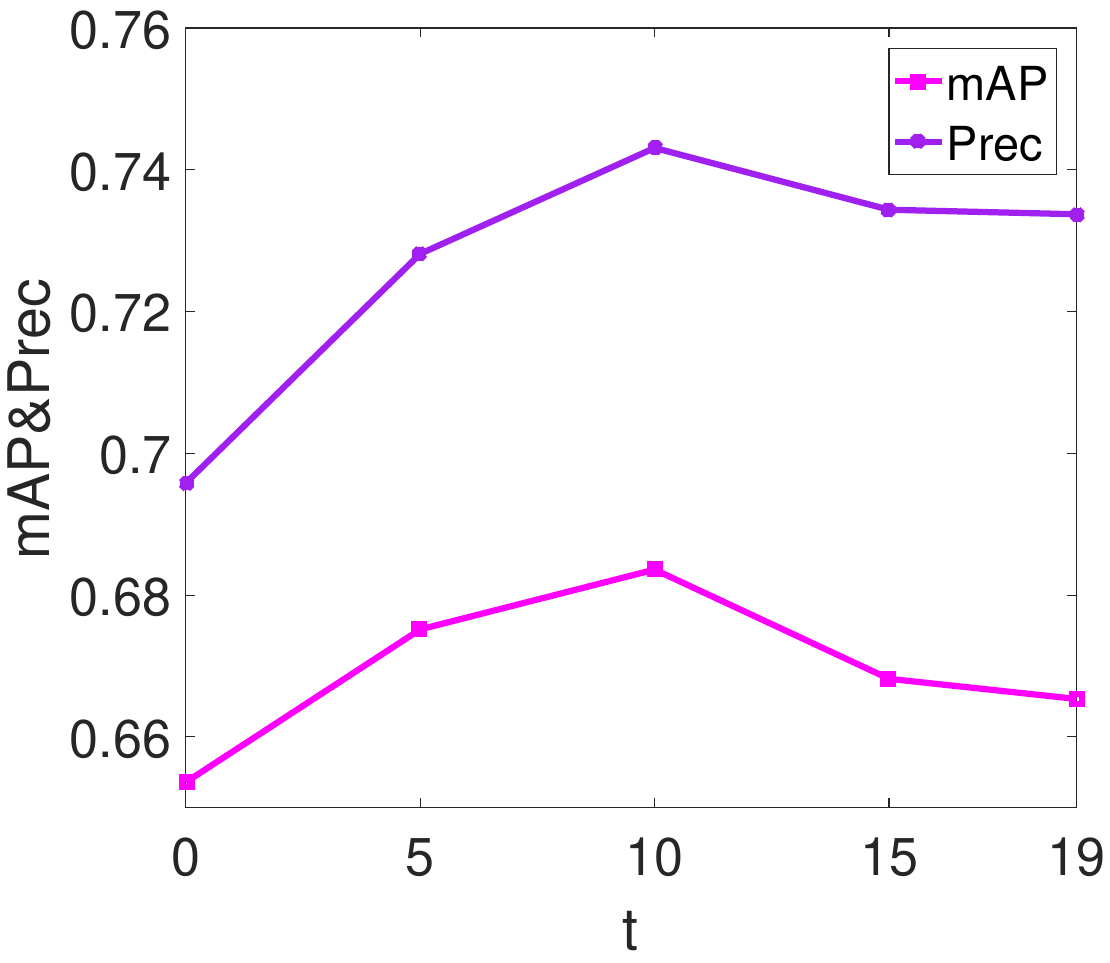}}
\subcaptionbox{\hspace{-0.7cm}\label{Fig.sub.3}}{
    \includegraphics[width=4cm,height=3cm]{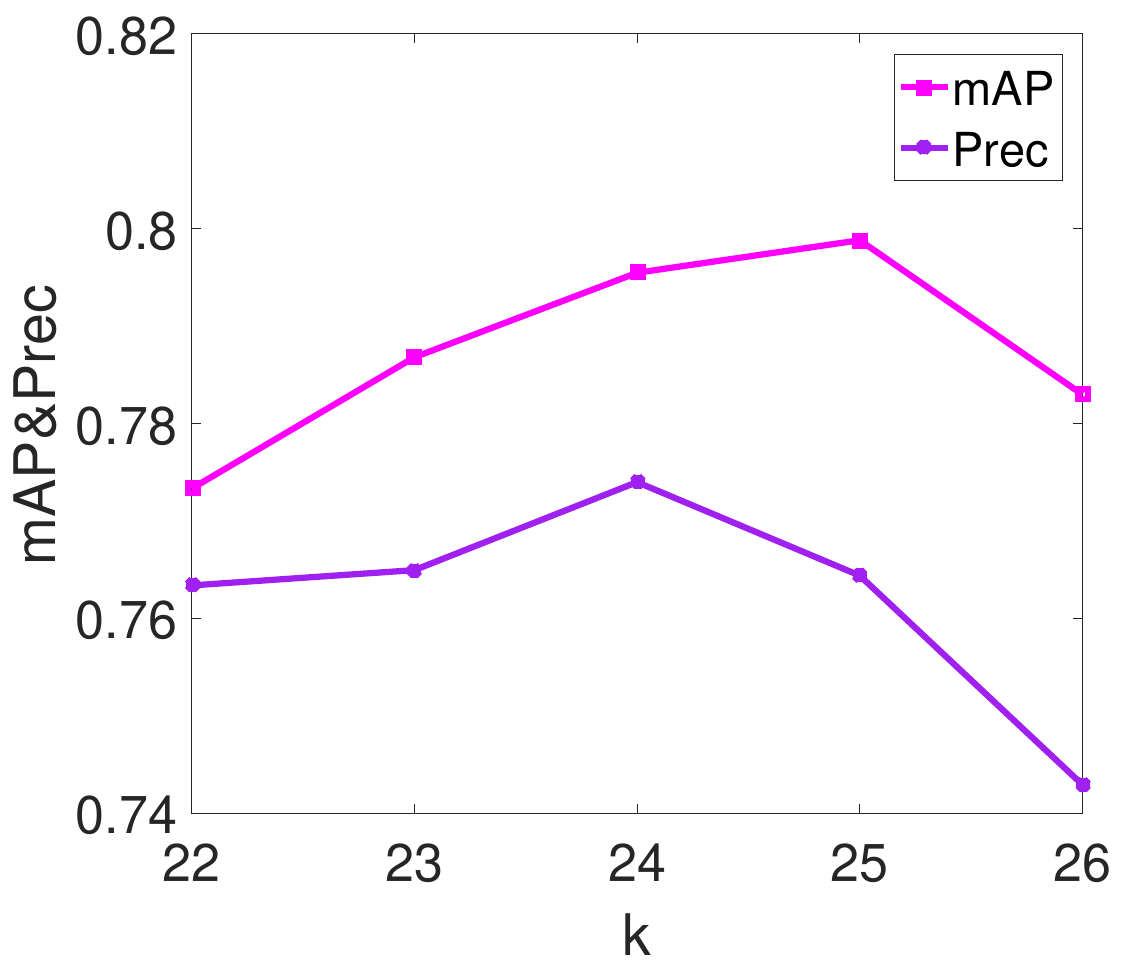}}
\caption{Parameter analysis results. (a) Varying the parameter $\lambda$. (b) Varying the parameter $t$. (c) Varying the parameter $k$.}
\label{fig3}
\end{figure*}

The proposed method SBKA achieves competitive performance on general public ZS-SBIR datasets. 
Benefiting from the designed knowledge alignment training and one-to-many matching strategies, the proposed method outperforms the SOTA methods by 8.6\% and 11.6\% with the mAP metric on the Sketchy Ext and TU-Berlin Ext datasets respectively. 
It is noted that the proposed method can achieve the best performance with the mAP metric, but exhibits slightly inferior performance compared with the SOTA methods with the precision metric.
This is because the mAP metric is calculated based on the precision-recall curve, which contains the precision values and a more comprehensive evaluation indicator.
Specifically, the proposed method also achieves superior performance with the precision metric on the Sketchy Ext Split dataset.
The experimental results prove the proposed algorithm can effectively reduce the adverse effects of the inter-modality gap through learning generalized discriminative representation.



\subsection{Parameters Analysis}
For the parameter $\lambda$ in Eq. \ref{equation5}, we choose values of 0.01, 0.1, 1 and 10 on the Sketchy Ext dataset to investigate their effects on model performance. 
To present it more visually, we took the logarithm with a base of 10 of the above four values, and the results are shown in Figure \ref{Fig.sub.1}. 
The parameter $\lambda$ balances the importance of the inter-modality alignment constraint and the bidirectional knowledge alignment constraints, which is important for retrieval performance.
As observed from Figure \ref{Fig.sub.1}, the model exhibits better performance when $\lambda$ is set to 0.1.

The parameter $t$ for symmetrical bidirectional knowledge alignment is also crucial for student model performance as shown in Algorithm 1. 
Starting bidirectional knowledge alignment too early can lead to model collapse, while starting it too late might not significantly improve model performance. 
We investigate the effect of $t$ on the Sketchy Ext dataset. 
After 20 training epochs, we initiate symmetrical bidirectional knowledge alignment respectively after the $0$-th, $5$-th, $10$-th, $15$-th, and $19$-th epochs following regular training. 
As shown in Figure \ref{Fig.sub.2}, we find that setting the parameter $t$ to approximately 10 leads to better performance.

In the one-to-many cluster matching stage, we evaluate the effect of the numbers of clusters $k$.
As shown in Figure \ref{Fig.sub.3} on the Sketchy Ext dataset, we find that the model achieves its optimum mAP metric when the cluster number $k$ is 25. It is worth noting that the number of clusters $k$ is the same as the number of unseen classes in the dataset.
In the testing stage, we opt to select several cluster numbers to adapt to different gallery sets.


\begin{figure*}[h]
\centering
\includegraphics[width=0.85\linewidth]{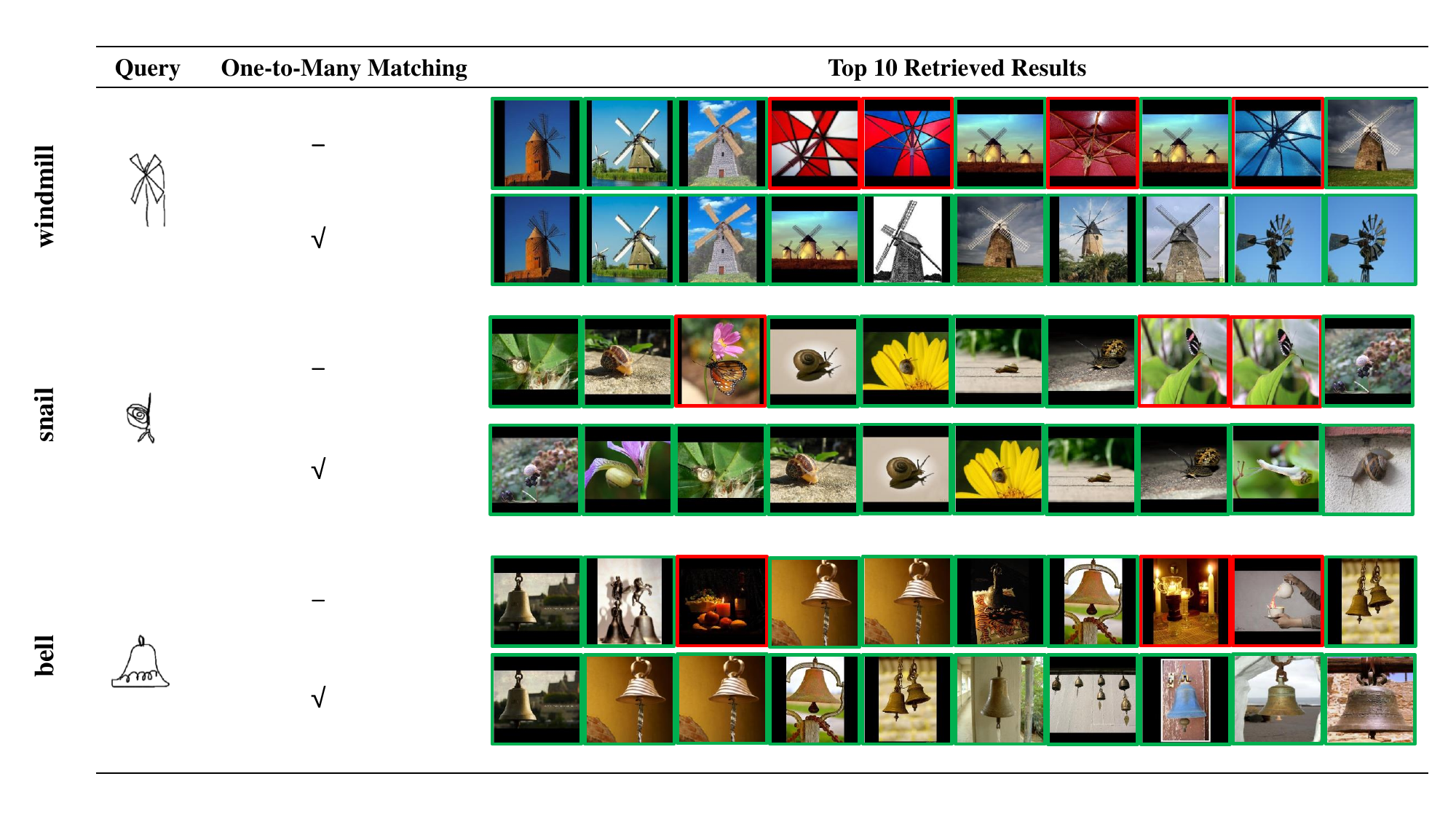} 
\caption{Exemplar comparison retrieval results for query sketches and the top 10 retrieved photos on Sketchy Ext Dataset. The green box denotes the correct result and the red box denotes the incorrect result.}
\label{fig2}
\end{figure*}

\subsection{Ablation Study}

The proposed method mainly contains two designed modules: the bidirectional knowledge alignment training module and the one-to-many matching module. To reveal how each module contributes to performance improvement, we conduct a comprehensive ablation study to analyze them
on the Sketchy Ext dataset and Sketchy Ext Split dataset as shown in Table \ref{table2}.

We utilize the unidirectional knowledge distillation algorithm as the baseline method for a fair comparison, which is constrained only by $L_{cls}$ and $L_{ma}$.
With the symmetrical bidirectional knowledge alignment component, the student model consistently outperforms the fixed knowledge distillation method. 
On the Sketchy Ext dataset, after adding this component, the mAP and precision evaluation metric of the proposed method increase by 3.8\% and 1.4\% respectively. 
This is because our designed symmetrical bidirectional knowledge alignment can learn robust modality-invariant discriminative knowledge through sufficient mutual interaction.
When additionally one-to-many cluster matching is utilized, the mAP and Prec metrics both increase, achieving the best retrieval performance.
Benefiting from the simple one-to-many cluster matching algorithm, which integrates the inherent relationships in the set and provides a simple way to decrease the intra-class variance.

\begin{table}[h]
\centering
\Large
\caption{Ablation study results on Sketchy Ext and Sketchy Ext Split datasets.}
\label{table2}
\resizebox{\columnwidth}{!}{
\begin{tabular}{cccccc}
    \toprule
    \multirow{2}{*}{\raisebox{-0.28cm}{\parbox{5cm}{\centering Bidirectional Knowledge Alignment}}} & \multirow{2}{*}{\raisebox{-0.28cm}{\parbox{3cm}{\centering One-to-Many Matching}}} & \multicolumn{2}{c}{Sketchy Ext} & \multicolumn{2}{c}{Sketchy Ext Split} \\
    \cmidrule{3-4} \cmidrule{5-6} 
    & & mAP@all & Prec@100 & mAP@200 & Prec@200 \\
    \midrule
     - & -  & 0.659 & 0.733 & 0.521 & 0.610  \\
     \checkmark & - & 0.683 & 0.743 & 0.532 & 0.619 \\
     \checkmark &\checkmark & \textbf{0.799} & \textbf{0.764} & \textbf{0.618} & \textbf{0.666} \\
    \bottomrule
\end{tabular}
}
\end{table}

\subsection{Algorithm Analysis}

\emph{Retrieval Results.}
Figure \ref{fig2} shows the retrieval results after we implement the one-to-many cluster matching method. 
For every sketch query, we show the top 10  retrieved results in the gallery. 
It can be found that the retrieval model is highly prone to failure when the photo contains geometrical characteristics similar to the query sketch.
For example, the geometrical contours of a windmill and an umbrella are similar, which can mislead the retrieval model. 
It is the key issue of the one-to-many cluster matching algorithm to improve retrieval performance, which sufficiently leverages the inherent relationship in the intra-class set.
After aggregating intra-class information, the retrieval results (shown in Figure \ref{fig2}) demonstrate that the proposed method can effectively distinguish between photos with similar contours but different classes.

\begin{figure}[h!]
\centering
\includegraphics[width=0.85\linewidth]{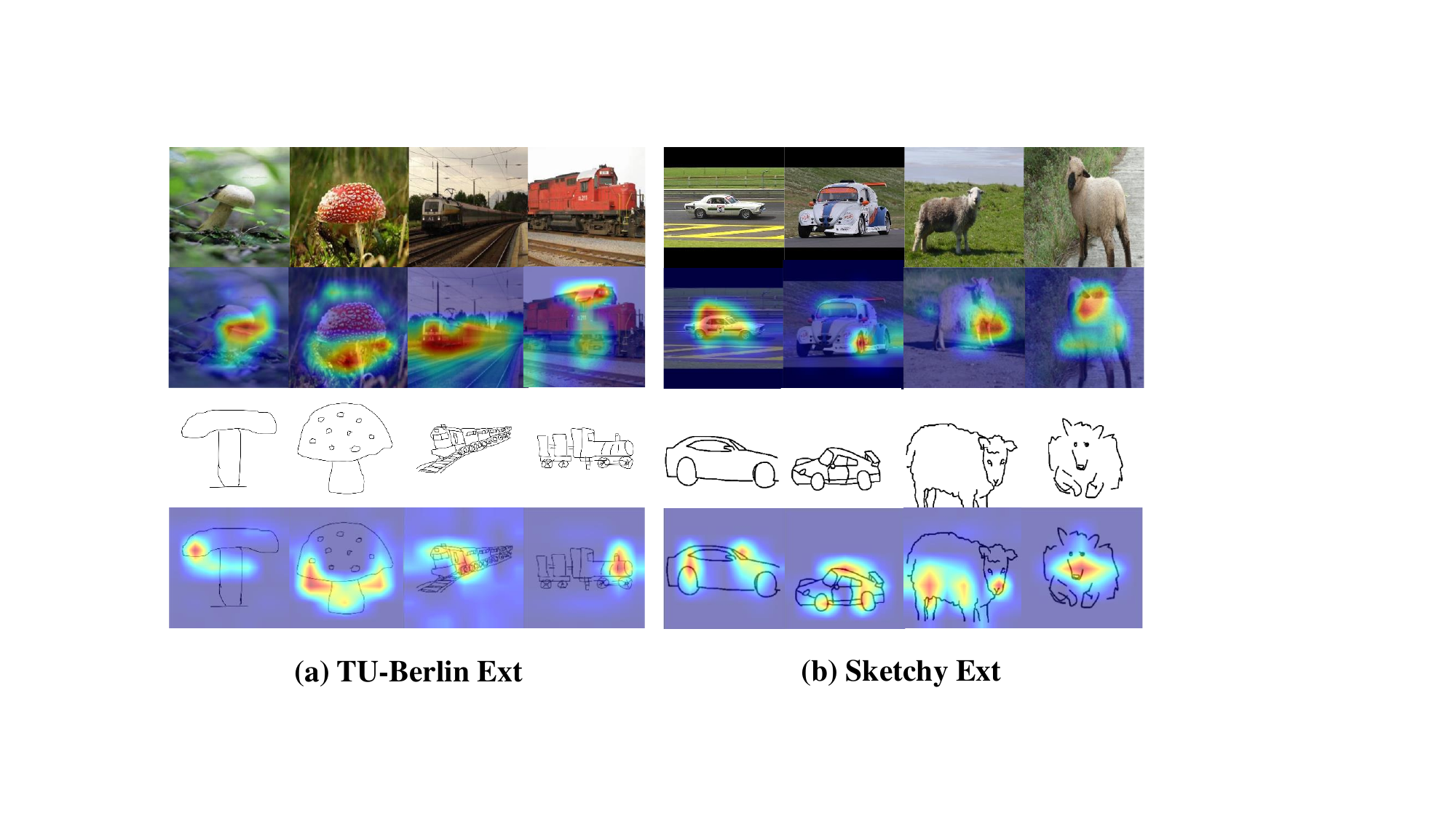} 
\caption{Grad-CAM visualization of samples on TU-Berlin Ext and Sketchy Ext datasets.}
\label{fig4}
\end{figure}

\emph{Visualization Analysis.}
To exhibit the discriminative information of sketch and photo features, we utilize Grad-CAM for feature visualization on the TU-Berlin and Sketchy Ext datasets. 
As shown in Figure \ref{fig4},  
we can find that the attention regions of the proposed representation model tend to be consistent across different modalities (e.g., sketches and photos). 
Additionally, a similar phenomenon can also be observed on the other ZR-SBIR datasets, which proves our bidirectional knowledge alignment method can efficiently extract discriminative information with strong generalizability. 
However, there exist some failure issues when the sketch contains a large pose variance compared with the gallery photo.
It also inspires researchers to focus on designing a strong discriminative classifier from the perspective of attention consistency.

\section{Conclusion}

\noindent 
In this paper, we propose a novel symmetrical bidirectional knowledge alignment method for zero-shot sketch-based image retrieval. 
The proposed method designs a bidirectional knowledge alignment learning strategy in the training stage, which will force the teacher and student models to learn generalized discriminative information for better knowledge alignment mutually.
Considering the large intra-class invariance in the ZS-SBIR task, we further propose a one-to-many cluster matching algorithm in the testing stage, which can effectively integrate the inherent relationship of intra-class gallery photos to boost retrieval performance.
Noting the proposed one-to-many cluster matching strategy is without complex network architecture.
Experiments on the public datasets illustrate the superior performance. 
In the future, we will evaluate the cross-modality matching performance on more complex real-world scenarios to design robust representations with good generalizability.

\clearpage
{
    \small
    \bibliographystyle{ieeenat_fullname}
    \bibliography{main}
}


\end{document}